\documentclass[10pt,conference]{IEEEtran}
\IEEEoverridecommandlockouts

\usepackage{comment}

\usepackage[numbers,sort&compress]{natbib}
\usepackage[hidelinks]{hyperref}
\usepackage{amsmath,amssymb,amsfonts}
\usepackage{textcomp}
\usepackage[table]{xcolor}
\usepackage{subcaption}
\usepackage{hyperref}
\usepackage{balance}
\usepackage{graphicx}
\usepackage{cleveref}
\usepackage{dblfloatfix}
\usepackage[font={small,it}]{caption}
\usepackage{tabularx}
\usepackage{float}
\usepackage{svg}
\usepackage{makecell}
\usepackage{enumitem}
\usepackage{pifont}
\usepackage{multirow}
\usepackage{booktabs}
\usepackage{arydshln}
\usepackage{colortbl}
\usepackage{diagbox}
\usepackage{longtable}
\usepackage[ruled,vlined]{algorithm2e}
\usepackage[cache=false]{minted}
\usepackage{microtype}
\definecolor{codebg}{RGB}{245,245,245}
\setminted{
  linenos,
  breaklines,
  breakanywhere,
  autogobble,
  fontsize=\footnotesize,
  bgcolor=codebg,
  frame=leftline,
  framesep=1.5mm,
  numbersep=8pt,           
  xleftmargin=2.6em,
  tabsize=4
}

\def\BibTeX{{\rm B\kern-.05em{\sc i\kern-.025em b}\kern-.08em
    T\kern-.1667em\lower.7ex\hbox{E}\kern-.125emX}}

\makeatletter

\newcommand*\titleheader[1]{\gdef\@titleheader{#1}}
\AtBeginDocument{%
  \let\st@red@title\@title
  \def\@title{%
    \bgroup\normalfont\large\centering\@titleheader\par\egroup
    \vskip1.5em\st@red@title}
}

\makeatother

\makeatletter
 \let\old@ps@headings\ps@headings
 \let\old@ps@IEEEtitlepagestyle\ps@IEEEtitlepagestyle
\def\confheader#1{%
  \def\ps@headings{%
    \old@ps@headings%
  }%
  \def\ps@IEEEtitlepagestyle{%
    \old@ps@IEEEtitlepagestyle%
    \def\@oddhead{%
      \begin{tabular}{@{}l@{}}
        \fontsize{8pt}{10pt}\selectfont\strut #1 \\ 
        \strut \hfill
      \end{tabular}%
    }%
    \def\@evenhead{%
      \begin{tabular}{@{}l@{}}
        \fontsize{8pt}{10pt}\selectfont\strut #1 \\ 
        \strut \hfill
      \end{tabular}%
    }%
  }%
  \ps@headings%
}

\makeatother

\begin{document}
\title{Learning to Verify: Signature-Guided Training for Adaptive Adversarial Detection}

\author{
Arash Vashagh and Roozbeh Razavi-Far\\
Trustworthy and Secure AI (TSAI) Lab,
Faculty of Computer Science,\\
University of New Brunswick, Fredericton, NB, Canada\\
arash.vashagh@unb.ca, roozbeh.razavi-far@unb.ca
}

\maketitle

\begin{abstract}
Adversarial example detectors often fail when attackers explicitly attempt to bypass them. We propose a dual-model approach that keeps the original classifier fixed and trains a separate verifier to assess its predictions. During training, we identify patterns of internal neural network activations that distinguish genuine examples of each class from adversarial examples targeting that class. These patterns, called activation signatures, guide the generation of challenging adversarial examples that mislead the classifier, challenge the verifier, and internally resemble the target class. The verifier learns to recognize these examples while preserving its performance on clean inputs through knowledge distillation. At inference, the verifier evaluates the classifier's prediction without requiring activation signatures. We evaluate the approach on publicly available datasets against several adaptive attacks, including attacks unseen during training. The proposed verifier substantially outperforms directly attacked confidence-based and feature-based detectors under the evaluated \(L_\infty\) attacks. Ablation studies show that adaptive verifier training is the main source of this improvement, while signature guidance primarily benefits clean accuracy. Additional \(L_2\) evaluations reveal limitations in robustness across perturbation types.
\end{abstract}


\section{Introduction}
\label{sec:introduction}

Deep neural networks achieve strong performance in image classification, but small input perturbations can cause confident but incorrect predictions \cite{7958570}. Adversarial training improves robustness by exposing a model to perturbed examples during optimization \cite{madry2017towards}, but its protection may depend on the attacks used during training, and retraining may reduce clean accuracy. Adversarial example detection provides an alternative by rejecting inputs that do not appear trustworthy. The problem becomes harder under adaptive attacks, where the attacker jointly optimizes misclassification and detector evasion \cite{10.1145/3128572.3140444,NEURIPS2020_11f38f8e}.

Existing detectors use classifier confidence, energy, feature-space distances, intermediate activations, or complementary prediction and rejection models \cite{hendrycks2017a,NEURIPS2020_f5496252,NEURIPS2018_abdeb6f5,metzen2017on,palumbo2023two}. However, fixed scores and simple model agreement can remain vulnerable when the attacker knows the detection rule. Detectors must also generalize beyond the attack procedures observed during training.

We study whether a separate verifier can provide an adaptive rejection signal without modifying the deployed classifier. The verifier should learn from attacks that challenge its decision rule while remaining simple at inference. A reliable evaluation should attack each detector directly, include unseen attacks, verify perturbation constraints, and report attack success.

We propose a signature-guided dual-model verifier in which a fixed classifier produces the prediction and a separately trained verifier measures support for that prediction. Our contribution combines detector-aware training of a separate verifier, a training-only signature curriculum, and prediction-conditioned rejection without a test-time signature bank. The signatures help construct examples that fool the classifier, challenge the verifier, and internally resemble the target class. The verifier learns to recover the original label while preserving clean behavior through label smoothing and distillation. During inference, the signature bank is removed, and detection uses the verifier's probability for the classifier's predicted class with a class-specific threshold. Deployment requires one forward pass through each model.

We evaluate the method on CIFAR-10, SVHN, and STL-10 using three random seeds. The evaluation includes detector-aware PGD and APGD, a CW-style $L_2$ attack, and AutoAttack candidates followed by detector-aware refinement. We also directly attack MSP, energy, diagonal Mahalanobis distance, and an activation-based logistic detector using their respective scores. The full verifier outperforms these baselines and behaves similarly across the three $L_\infty$ procedures, including APGD and AA+Refine, which are unseen during the verifier training. AA+Refine generates adversarial candidates using AutoAttack \cite{pmlr-v119-croce20b}, and then, applies detector-aware PGD refinement. We further compare a no-signature verifier and evaluate DenseNet121 and direct $L_2$ PGD. These comparisons distinguish the value of adaptive verifier training from the measured effect of signature guidance on clean predictions. The implementation is available at \url{https://github.com/arashVsh/learning-to-verify-ssci-2027}.

\section{Background}

\label{sec:background}

This section reviews robust learning, adversarial detection, rejection, and verification methods, as well as adaptive evaluation. Recent surveys provide a broader view of adversarial machine learning by organizing attacks according to attacker knowledge, objectives, capabilities, and learning stage, and by grouping defenses into several strategies \cite{arashSurvey,202607.1022}. Within this broader landscape, our work focuses on test-time evasion detection and on evaluation against adaptive white-box attacks.

\subsection{Adversarial Examples and Robust Learning}

Adversarial training minimizes loss under perturbations generated within a predefined threat model \cite{madry2017towards}. Its protection may weaken when the test-time attack differs from the training attack, which limits conclusions drawn from a single perturbation model. Continual robust training addresses evolving adversaries without repeatedly rebuilding the model \cite{pmlr-v267-dai25c}; more broadly, self-adaptive continual learning addresses harsh, changing conditions \cite{RAZAVIFAR2026134136}. Decision-based methods further demonstrate that adversarial examples can be produced without access to gradients or model parameters \cite{ICLR2026_c81690e2}. This broad range of attack settings shows that robustness must be evaluated beyond one training algorithm or access assumption.

\subsection{Adversarial Example Detection}

Early detection methods train auxiliary networks using intermediate features, while later robust objectives incorporate adversarial examples into detector training \cite{metzen2017on,NEURIPS2018_e0f7a4d0}. Other approaches use class-conditional feature distances, unusual activation subsets, neural fingerprints, or fuzzy rules \cite{NEURIPS2018_abdeb6f5,speakman2018subset,fisher2024neural,10372061}. These methods differ in how they represent abnormal behavior, but most depend on a score learned or calculated from the classifier's internal response. Layer-wise responses and graph-guided model behavior provide additional signals by capturing where perturbations affect the network and how model behaviors relate \cite{mumcu2025universal,CHEN2024103710}. Robust anomaly detection and sequential statistical testing instead address weaknesses in learned negative examples and violations of exchangeability \cite{ICLR2025_4fa8c1f3,pmlr-v266-dong25a}. 
Recent work has also explored transferring adversarial detectors across different classifier architectures. GraphRectify uses graph-based representations of intermediate features to adapt detectors between neural network backbones \cite{vashagh2026graphrectify}.

Output-based scores remain useful baselines because they require no additional feature extractors. MSP identifies unusually weak classifier confidence, while energy uses the complete logit vector rather than only its largest probability \cite{hendrycks2017a,NEURIPS2020_f5496252}. Energy-based modeling has also been connected to scalable adversarial training \cite{ICLR2026_6fcb1afc}. These inexpensive scores may nevertheless miss adversarial inputs that receive confident incorrect predictions.

\subsection{Verification, Rejection, and Generalization}

Rejecting an input can be preferable when the evidence for a prediction is insufficient. Two-head models combine prediction and rejection, agreement checking compares model behavior, and reject-option learning balances clean accuracy, robustness, and abstention \cite{palumbo2023two,nirala2025alignfix,shah2024towards}. Selective-classification theory further studies when abstention is preferable to a potentially incorrect decision \cite{ICLR2026_3fe2a777}. In this setting, a model is allowed to withhold a prediction on uncertain inputs, so performance is considered jointly in terms of prediction quality and the fraction of inputs on which the model chooses to decide.

Detectors must also recognize attacks not observed during the training session. Few-shot approaches use limited attack examples to reduce reliance on large labeled collections \cite{liu2024enhancing}. Distillation offers another way to transfer predictive structure, with robust variants encouraging a student to reproduce useful teacher behavior under perturbations \cite{lee2025indirect,ICLR2026_f68548dd}. These studies motivate verification systems that use complementary model behavior rather than one confidence estimate.

\subsection{Adaptive Evaluation}

A white-box attacker has access to the detector's decision rule and can optimize classification success and detector evasion jointly \cite{10.1145/3128572.3140444,NEURIPS2020_11f38f8e}. Detectors may appear effective against fixed classifier attacks yet fail once their own score is included in the objective. Attacks that cannot optimize through a defense may also create misleading robustness through gradient masking \cite{pmlr-v80-athalye18a}. Diverse attack ensembles and diagnostic indicators help to distinguish genuine robustness from failed optimization \cite{pmlr-v119-croce20b,NEURIPS2022_91ffdc5e}. Detection can remain as difficult as robust classification under a fully informed adversary, and test-time defenses require evaluation of their complete adaptive behavior \cite{pmlr-v162-tramer22a,pmlr-v162-croce22a}. A credible evaluation should therefore attack each detector directly, verify perturbation limits, and report attack success so the detector's performance is not attributed to weak attacks.

\section{Methodology}
\label{sec:method}

\begin{figure*}[t]
\centering
\includegraphics[width=\textwidth]{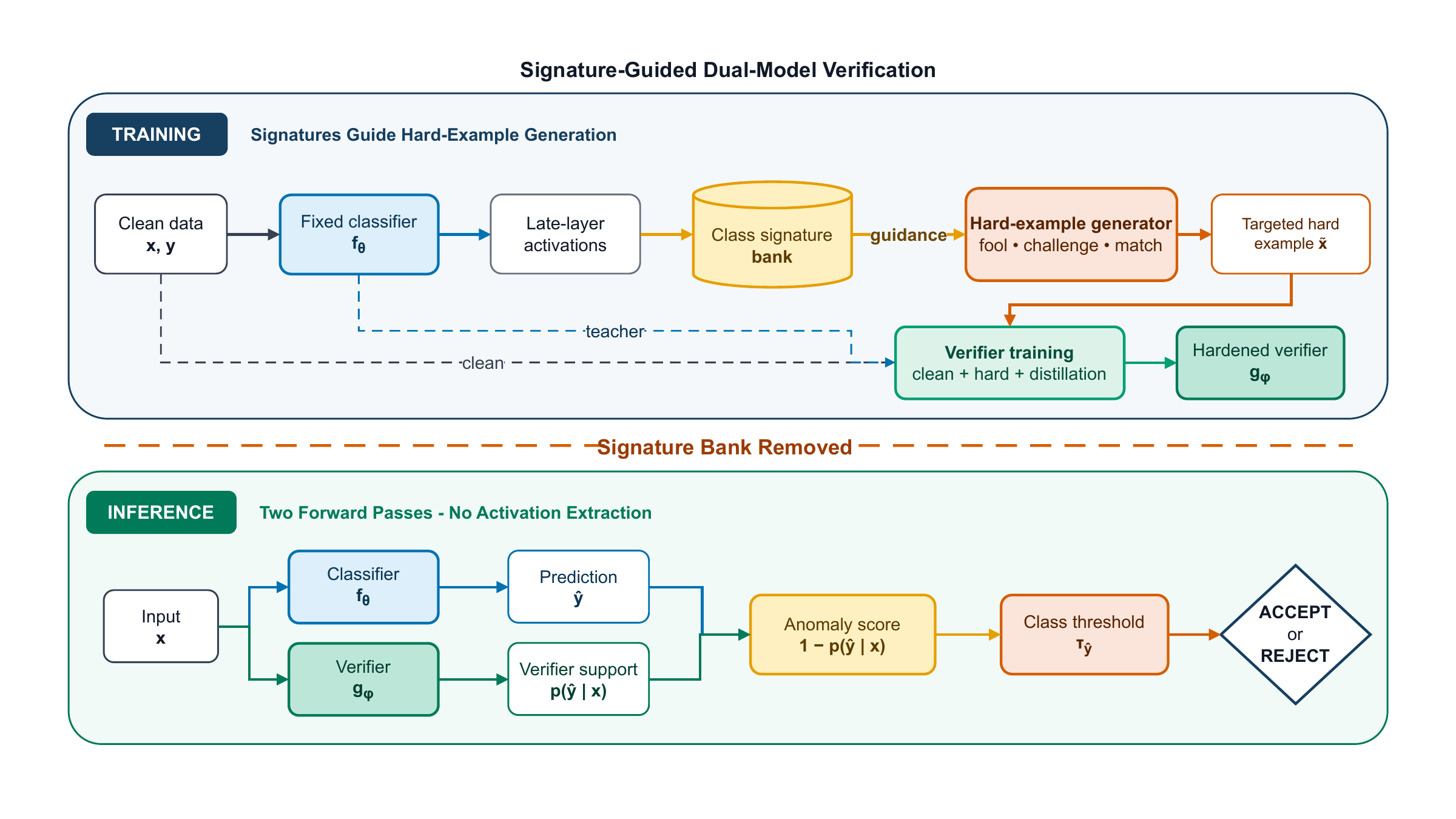}
\caption{Overview of the proposed method. Activation signatures guide
hard-example generation during the training session. During inference, the fixed
classifier and hardened verifier provide the prediction and verification
score used by the class-specific rejection rule.}
\label{fig:method_overview}
\end{figure*}

This section describes signature construction, signature-guided hard-example
generation, verifier training, and the final prediction-conditioned
rejection rule. Figure~\ref{fig:method_overview} summarizes the proposed
verifier.

Let $\mathcal{C}={1,\ldots,C}$ denote the set of $C$ classes. The labeled training set is divided into four disjoint partitions. The signature, verifier, development, and calibration partitions are denoted by $\mathcal{D}_{\mathrm{sig}}$, $\mathcal{D}_{\mathrm{ver}}$, $\mathcal{D}_{\mathrm{dev}}$, and $\mathcal{D}_{\mathrm{cal}}$, respectively. They contain 20\%, 50\%, 15\%, and 15\% of the training samples with class-level stratification. The base classifier is trained using $\mathcal{D}_{\mathrm{sig}}\cup\mathcal{D}_{\mathrm{ver}}$. Signature construction uses $\mathcal{D}_{\mathrm{sig}}$. Verifier optimization uses $\mathcal{D}_{\mathrm{ver}}$, while checkpoint selection and threshold calibration use $\mathcal{D}_{\mathrm{dev}}$ and $\mathcal{D}_{\mathrm{cal}}$. The official test set $\mathcal{D}_{\mathrm{test}}$ is reserved for final evaluation.

The fixed classifier is denoted by $f_{\boldsymbol{\theta}}$. Its parameters are represented by $\boldsymbol{\theta}$. The verifier is denoted by $g_{\boldsymbol{\phi}}$. Its trainable parameters are represented by $\boldsymbol{\phi}$. Both models return vectors of logits. The classifier probability distribution is
$\pi_{\boldsymbol{\theta}}(\cdot\mid\mathbf{x})
=\operatorname{softmax}(f_{\boldsymbol{\theta}}(\mathbf{x}))$.
The verifier probability distribution is
$\pi_{\boldsymbol{\phi}}(\cdot\mid\mathbf{x})
=\operatorname{softmax}(g_{\boldsymbol{\phi}}(\mathbf{x}))$.
The input is denoted by $\mathbf{x}$ and its true label is $y\in\mathcal{C}$.

\subsection{Signature-Guided Verifier Training}

We construct one dataset-level signature bank with a class-conditioned entry for every class. Let $\mathbf{a}(\mathbf{x})$ denote the activation vector formed by flattening and concatenating outputs from selected late layers of $f_{\boldsymbol{\theta}}$. The implementation extracts activations from the last eight layers whose parameter counts exceed a fixed significance threshold.

Clean activations are collected from $\mathcal{D}_{\mathrm{sig}}$. Targeted $L_\infty$ PGD examples are also generated, but an example is retained only when $f_{\boldsymbol{\theta}}$ predicts its predetermined target. Target labels are selected before optimization and balanced across classes. If the finite guidance partition cannot provide the requested quota, the largest available balanced pool is retained as long as every target reaches at least half of that quota.

For each class $c\in\mathcal{C}$, mutual information identifies coordinates that distinguish genuine examples of class $c$ from successful adversarial examples targeting $c$. A variance filter first removes coordinates with little variation. The implementation retains at most 8192 variance candidates, and, then, selects at most 256 coordinates for each class.

Let $\mathcal{I}_c$ denote the set of $K$ coordinates selected for class $c$. For coordinate $j\in\mathcal{I}_c$, the median clean activation is $\mu_{cj}$. Its robust scale is $\rho_{cj}$, which is calculated as the median absolute deviation multiplied by 1.4826. A positive floor is applied to $\rho_{cj}$ to prevent unstable normalization.

The smooth $L_1$ function is denoted by $h$. Its quadratic branch is $h(q)=q^2/2$ for $|q|<1$. Its linear branch is $h(q)=|q|-1/2$ for $|q|\geq1$. The robust distance between $\mathbf{x}$ and the signature of class $c$ is

\begin{equation}
d_c(\mathbf{x})=
\frac{1}{K}\sum_{j\in\mathcal{I}_c}
h\left(
\frac{a_j(\mathbf{x})-\mu_{cj}}{\rho_{cj}}
\right),
\label{eq:signature_distance}
\end{equation}
where $a_j(\mathbf{x})$ is coordinate $j$ of the activation vector. A smaller distance indicates greater similarity to the clean activation pattern of class $c$.

An input may be close to several class signatures. Therefore, the selected class is compared with every competing entry in the bank. Let $\gamma>0$ denote the soft-minimum temperature. The competing-class distance is

\begin{equation}
o_c(\mathbf{x})=
-\gamma\log
\sum_{k\in\mathcal{C},\,k\ne c}
\exp\left(-\frac{d_k(\mathbf{x})}{\gamma}\right).
\label{eq:competing_distance}
\end{equation}

Intuitively, the soft minimum emphasizes the nearest competing class while retaining gradients from the others. The relative signature margin is $m_c(\mathbf{x})=
d_c(\mathbf{x})-o_c(\mathbf{x})$.

A negative margin indicates that the input is closer to class $c$ than to its strongest competitor. Distance and margin are standardized before they are combined. The resulting cross-bank geometry penalty is

\begin{equation}
\begin{aligned}
G(\mathbf{x},c)= h\left(\frac{d_c(\mathbf{x})-\bar d_c}{s_c^d}\right)\
+h\left(\frac{m_c(\mathbf{x})-\bar m_c}{s_c^m}\right),
\end{aligned}
\label{eq:cross_bank_geometry}
\end{equation}
where $\bar d_c$ and $s_c^d$ are the robust location and scale of distances for clean guidance inputs predicted as class $c$. The corresponding robust location and scale of the margins are $\bar m_c$ and $s_c^m$. This standardization prevents either component from dominating the other one due its numerical range. The two standardized terms are then added with equal weight so that hard-example generation penalizes both poor target-class proximity and poor separation from competing classes; standardization makes this unweighted combination comparable across the two quantities.

The verifier is initialized as a copy of the trained base classifier, so both models begin with the same clean decision structure. The base classifier is then kept fixed, while only the verifier is updated. To avoid adding target-sampling randomness to verifier training, a target label $r\in\mathcal{C}\setminus\{y\}$ is selected deterministically during each update. Let
$\mathcal{B}_{\infty}(\mathbf{x},\varepsilon)
=\{\mathbf{u}:\lVert\mathbf{u}-\mathbf{x}\rVert_{\infty}\leq\varepsilon\}$
denote the permitted perturbation region. Pixel values are also clipped to the valid input range. A hard training example $\widetilde{\mathbf{x}}$ is approximated by

\begin{equation}
\widetilde{\mathbf{x}}\approx
\underset{\mathbf{u}\in\mathcal{B}_{\infty}(\mathbf{x},\varepsilon)}
{\arg\min}
\Big[
\ell_{\mathrm{ce}}(f_{\boldsymbol{\theta}}(\mathbf{u}),r)
-\lambda_{\mathrm{v}}\ell_{\mathrm{ce}}(g_{\boldsymbol{\phi}}(\mathbf{u}),y)
+\lambda_{\mathrm{s}}G(\mathbf{u},r)
\Big],
\label{eq:verifier_training_attack}
\end{equation}
where $\ell_{\mathrm{ce}}$ denotes standard cross-entropy loss. The coefficient $\lambda_{\mathrm{v}}$ controls the pressure placed on the current verifier. The coefficient $\lambda_{\mathrm{s}}$ determines the strength of signature camouflage.

The first term moves the base classifier toward the target label. The negative second term searches for inputs on which the verifier loses the true label. The last term encourages the adversarial representation to resemble a clean member of the target class. We use an $L_\infty$ radius of $\varepsilon=8/255$. The attack performs seven PGD iterations with a step size of $2/255$. The verifier coefficient is set to $\lambda_{\mathrm{v}}=1$. The signature coefficient is set to $\lambda_{\mathrm{s}}=0.05$.

The verifier learns to recover the original label from these hard examples while retaining its clean predictive behavior. Let $\mathbb{E}_{\mathrm{ver}}$ denote expectation over labeled samples from $\mathcal{D}_{\mathrm{ver}}$. The verifier objective is

\begin{equation}
\begin{aligned}
\mathcal{L}_{\mathrm{ver}}(\boldsymbol{\phi})
={}&\mathbb{E}_{\mathrm{ver}}\Big[
\ell_{\mathrm{ls}}(g_{\boldsymbol{\phi}}(\mathbf{x}),y)
+\alpha\ell_{\mathrm{ls}}(g_{\boldsymbol{\phi}}(\widetilde{\mathbf{x}}),y)\\
&+\beta D_{\mathrm{KL}}\Big(
\pi_{\boldsymbol{\theta}}(\cdot\mid\mathbf{x})
\Vert
\pi_{\boldsymbol{\phi}}(\cdot\mid\mathbf{x})
\Big)
\Big].
\end{aligned}
\end{equation}

Here, $\ell_{\mathrm{ls}}$ denotes cross-entropy with label smoothing of 0.05. Adversarial recovery is weighted by $\alpha=1.5$. Clean distillation is weighted by $\beta=0.5$, and $D_{\mathrm{KL}}$ denotes KL divergence.

The parameters $\boldsymbol{\theta}$ remain fixed during the verifier training. After every epoch, the verifier is evaluated on $\mathcal{D}_{\mathrm{dev}}$. The retained checkpoint maximizes the average of clean accuracy and robust accuracy under a direct joint attack.

\subsection{Prediction-Conditioned Verification}

The signature bank and activation extractor are removed during inference. The base classifier first outputs the predicted class

\begin{equation}
\widehat{y}(\mathbf{x})=
\underset{c\in\mathcal{C}}{\arg\max},
f_{\boldsymbol{\theta},c}(\mathbf{x}),
\label{eq:base_prediction}
\end{equation}
where $f_{\boldsymbol{\theta},c}(\mathbf{x})$ is the logit assigned to class $c$. The detector then measures how strongly the verifier supports this prediction. Its anomaly score is

\begin{equation}
A(\mathbf{x})=
1-\pi_{\boldsymbol{\phi}}
\left(\widehat{y}(\mathbf{x})\mid\mathbf{x}\right).
\label{eq:anomaly_score}
\end{equation}

A high value of $A(\mathbf{x})$ indicates that the verifier assigns little probability to the class selected by the base classifier. This confidence-sensitive score is more informative than checking only whether the models produce different top-1 labels.

A clean calibration input is denoted by $\mathbf{z}$. For class
$c$, let $\mathcal{D}_{\mathrm{cal}}(c)$ contain the calibration
inputs whose base prediction is $c$. The class-specific threshold is

\begin{equation}
\tau_c=
\operatorname{Quantile}_{0.95}
\left(
A(\mathbf{z})\mid
\mathbf{z}\in\mathcal{D}_{\mathrm{cal}}(c)
\right).
\label{eq:class_threshold}
\end{equation}

If fewer than 20 inputs belong to $\mathcal{D}_{\mathrm{cal}}(c)$,
the global 95th-percentile clean threshold is used instead.

The final rejection decision is

\begin{equation}
R(\mathbf{x})=
\mathbb{I}\left[
A(\mathbf{x})\geq
\tau_{\widehat{y}(\mathbf{x})}
\right],
\label{eq:rejection_rule}
\end{equation}
where $R(\mathbf{x})=1$ indicates that the input is rejected. Class-conditioned calibration accommodates differences among predicted classes while maintaining an empirical clean false-positive rate near 5\%. Deployment requires one forward pass through each model and does not depend on activation extraction or signature comparison.

\section{Results}
\label{sec:results}
This section presents the experimental protocol and evaluates clean
performance, robustness to adaptive and unseen attacks, baseline
comparisons, and component ablations.

\subsection{Experimental Protocol}
\begin{table}[t]
\caption{Clean performance across three seeds. Values are reported as mean $\pm$ standard deviation.}
\label{tab:clean_results}
\centering
\small
\setlength{\tabcolsep}{2.5pt}
\begin{tabular}{lccc}
\hline
Metric & CIFAR-10 & SVHN & STL-10 \\
\hline
Official base acc. (\%) &
$92.98{\pm}0.32$ &
$96.74{\pm}0.10$ &
$74.26{\pm}2.40$ \\

Balanced base acc. (\%) &
$93.43{\pm}0.48$ &
$96.13{\pm}0.45$ &
$73.37{\pm}2.33$ \\

Verifier acc. (\%) &
$85.70{\pm}1.06$ &
$93.17{\pm}0.23$ &
$68.30{\pm}5.20$ \\

Clean FPR (\%) &
$5.55{\pm}0.80$ &
$5.13{\pm}0.33$ &
$5.32{\pm}0.93$ \\

Latency (ms/sample) &
$0.359{\pm}0.009$ &
$0.375{\pm}0.131$ &
$1.405{\pm}0.005$ \\
\hline
\end{tabular}
\end{table}

\begin{table*}[t]
\caption{Detector performance under direct adaptive attacks. All entries are percentages reported as mean $\pm$ standard deviation over three seeds. PGD is seen during the verifier training. The remaining attacks are unseen.}
\label{tab:adaptive_results}
\centering
\scriptsize
\setlength{\tabcolsep}{10pt}
\begin{tabular}{llcccc}
\hline
Dataset & Attack & AUC $\uparrow$ &
$\mathrm{TPR}_{1\%}$ $\uparrow$ &
$\mathrm{TPR}_{5\%}$ $\uparrow$ &
Bypass $\downarrow$ \\
\hline
CIFAR-10 & PGD &
$79.10{\pm}2.22$ & $21.53{\pm}2.72$ &
$38.83{\pm}5.58$ & $57.23{\pm}3.04$ \\

CIFAR-10 & APGD &
$79.06{\pm}2.21$ & $21.53{\pm}2.46$ &
$38.57{\pm}5.30$ & $57.50{\pm}2.95$ \\

CIFAR-10 & AA+Refine &
$79.16{\pm}2.19$ & $21.53{\pm}2.60$ &
$39.00{\pm}5.43$ & $57.10{\pm}2.93$ \\

CIFAR-10 & CW $L_2$ &
$64.54{\pm}1.39$ & $1.43{\pm}0.35$ &
$2.73{\pm}0.42$ & $95.63{\pm}1.17$ \\
\hline
SVHN & PGD &
$80.80{\pm}0.95$ & $24.80{\pm}0.53$ &
$41.73{\pm}2.66$ & $56.90{\pm}1.42$ \\

SVHN & APGD &
$80.66{\pm}0.89$ & $24.80{\pm}0.30$ &
$42.13{\pm}1.70$ & $56.80{\pm}1.28$ \\

SVHN & AA+Refine &
$81.17{\pm}0.80$ & $25.83{\pm}0.12$ &
$43.00{\pm}2.01$ & $55.97{\pm}1.55$ \\

SVHN & CW $L_2$ &
$71.51{\pm}1.69$ & $1.60{\pm}0.10$ &
$2.37{\pm}0.06$ & $94.53{\pm}1.15$ \\
\hline
STL-10 & PGD &
$79.86{\pm}1.74$ & $28.97{\pm}3.41$ &
$46.90{\pm}1.77$ & $51.30{\pm}3.56$ \\

STL-10 & APGD &
$79.92{\pm}1.71$ & $29.40{\pm}3.41$ &
$47.03{\pm}1.66$ & $51.20{\pm}3.58$ \\

STL-10 & AA+Refine &
$79.87{\pm}1.74$ & $28.87{\pm}3.41$ &
$46.87{\pm}1.72$ & $51.33{\pm}3.50$ \\

STL-10 & CW $L_2$ &
$87.87{\pm}1.70$ & $47.23{\pm}1.31$ &
$62.00{\pm}2.35$ & $37.60{\pm}1.61$ \\
\hline
\end{tabular}
\end{table*}

The main experiments use ResNet-18 and seeds 0, 1, and 2. An additional CIFAR-10 experiment uses DenseNet121 with seed 0. Main results are reported as the mean and sample standard deviation across three seeds. The signature soft-minimum temperature is $\gamma=0.25$, and eligible layers contain more than 1,000 parameters. 
The classifier and verifier use stochastic gradient descent (SGD) for 50 and 20 epochs at initial learning rates of 0.1 and 0.01, respectively, with momentum 0.9, weight decay $5\times10^{-4}$, Nesterov momentum, and cosine decay. Batch sizes are reduced when needed to fit GPU memory. Each adaptive evaluation retains 1,000 successful targeted adversarial examples, with 100 examples assigned to each target class. Only test inputs classified correctly before the attack are eligible.

PGD, APGD, and the AutoAttack candidate refinement use an $L_\infty$ limit of $8/255$. The CW-style attack uses an $L_2$ limit of 3.0. PGD is the only attack family used during the verifier training. Detector-aware PGD and APGD use 80 iterations; CW uses 160 iterations at a learning rate of 0.01. AA+Refine uses 80 detector-aware PGD refinement iterations. The additional direct $L_2$ PGD evaluation uses radius 3.0. APGD, CW, and AutoAttack candidate refinement are unseen during the training session.

We report area under the receiver operating characteristic curve, true-positive rate at 1\% and 5\% false-positive rates, and detector bypass rate. The bypass rate is the fraction of successful targeted adversarial examples accepted by the detector. For every retained adversarial example, we verified target-class success and compliance with the specified $L_\infty$ or $L_2$ constraint. Across the evaluated configurations, targeted attack success ranged from 96.4\% to approximately 100\%, with no norm violations.

Table~\ref{tab:clean_results} summarizes clean behavior. Official base accuracy is measured over the complete test set, while balanced base and verifier accuracies use the same class-balanced set of 2,000 clean test inputs used for detector evaluation. The calibrated rejection rule yields a clean FPR between 5.13\% and 5.55\%.

\subsection{Adaptive Detection and Unseen Attacks}

Table~\ref{tab:adaptive_results} reports similar detection performance under PGD, APGD, and AA+Refine across all three datasets. These results cover both the PGD procedure used during verifier training and two additional $L_\infty$ evaluation procedures. At the calibrated 5\% false-positive rate (FPR), the detector identifies a substantial portion of successful attacks, with the highest detection rate on STL-10.

The CW-style $L_2$ evaluation produces a different dataset-dependent pattern, with stronger detection on STL-10 than on CIFAR-10 or SVHN. Across the evaluated configurations, targeted attack success is high and all retained examples satisfy their perturbation constraints. The implications of these differences across attack procedures and perturbation norms are discussed in Section~\ref{sec:discussion}.

\subsection{Baseline Comparison and Ablation}

Each baseline is evaluated using a direct adaptive attack optimized against its own anomaly score. Table~\ref{tab:baseline_results} compares these results with the proposed detector under direct joint PGD.

\begin{table}[H]
\caption{Detector-aware $L_\infty$ attack results (mean $\pm$ standard deviation, percent). Each attack targets the corresponding detector score.}
\label{tab:baseline_results}
\centering
\scriptsize
\setlength{\tabcolsep}{2pt}
\resizebox{\columnwidth}{!}{%
\begin{tabular}{llccc}
\hline
Dataset & Detector & AUC $\uparrow$ &
$\mathrm{TPR}_{5\%}$ $\uparrow$ &
Bypass $\downarrow$ \\
\hline
CIFAR-10 & Proposed &
$\mathbf{79.10{\pm}2.22}$ &
$\mathbf{38.83{\pm}5.58}$ &
$\mathbf{57.23{\pm}3.04}$ \\

CIFAR-10 & Activation logistic &
$3.03{\pm}1.15$ & $0.00{\pm}0.00$ & $98.00{\pm}0.82$ \\

CIFAR-10 & Diagonal Mahalanobis &
$0.55{\pm}0.11$ & $0.00{\pm}0.00$ & $100.00{\pm}0.00$ \\

CIFAR-10 & Energy &
$1.92{\pm}1.24$ & $0.60{\pm}0.61$ & $99.27{\pm}0.68$ \\

CIFAR-10 & MSP &
$2.72{\pm}1.28$ & $0.50{\pm}0.53$ & $99.37{\pm}0.61$ \\
\hline
SVHN & Proposed &
$\mathbf{80.80{\pm}0.95}$ &
$\mathbf{41.73{\pm}2.66}$ &
$\mathbf{56.90{\pm}1.42}$ \\

SVHN & Activation logistic &
$8.26{\pm}1.33$ & $0.00{\pm}0.00$ & $98.60{\pm}0.56$ \\

SVHN & Diagonal Mahalanobis &
$16.17{\pm}0.56$ & $1.43{\pm}0.38$ & $96.97{\pm}0.86$ \\

SVHN & Energy &
$7.00{\pm}0.15$ & $2.37{\pm}0.32$ & $97.00{\pm}0.35$ \\

SVHN & MSP &
$9.10{\pm}1.10$ & $2.27{\pm}0.51$ & $97.17{\pm}0.95$ \\
\hline
STL-10 & Proposed &
$\mathbf{79.86{\pm}1.74}$ &
$\mathbf{46.90{\pm}1.77}$ &
$\mathbf{51.30{\pm}3.56}$ \\

STL-10 & Activation logistic &
$36.96{\pm}2.85$ & $0.00{\pm}0.00$ & $100.00{\pm}0.00$ \\

STL-10 & Diagonal Mahalanobis &
$11.83{\pm}2.65$ & $0.17{\pm}0.15$ & $99.70{\pm}0.17$ \\

STL-10 & Energy &
$1.16{\pm}1.22$ & $0.07{\pm}0.12$ & $99.93{\pm}0.12$ \\

STL-10 & MSP &
$1.63{\pm}1.44$ & $0.03{\pm}0.06$ & $99.90{\pm}0.17$ \\
\hline
\end{tabular}%
}
\end{table}

\begin{figure*}[!t]
\centering

\begin{subfigure}[t]{0.49\textwidth}
    \centering
    \includegraphics[width=\linewidth]{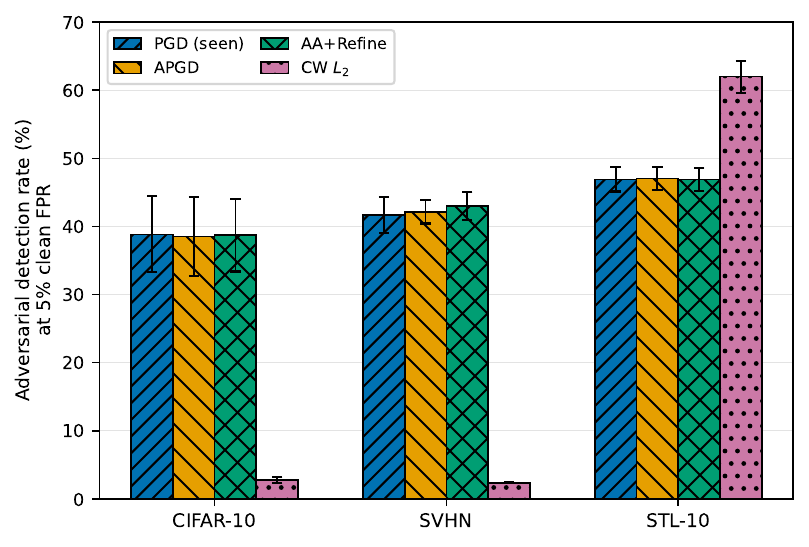}
    \caption{Adversarial detection rate across adaptive attacks when the detection threshold is calibrated to incorrectly reject 5\% of clean inputs.}
\end{subfigure}
\hfill
\begin{subfigure}[t]{0.49\textwidth}
    \centering
    \includegraphics[width=\linewidth]{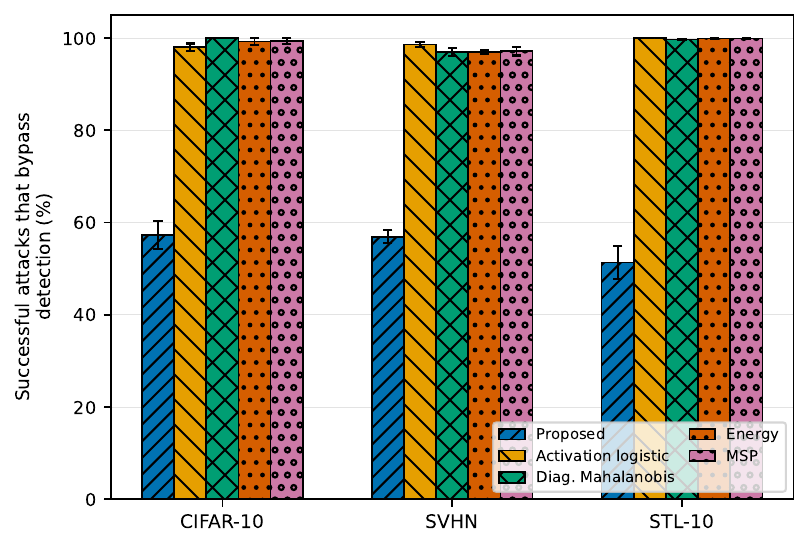}
    \caption{Detector bypass under direct adaptive attacks.}
\end{subfigure}

\caption{Adaptive detection results across CIFAR-10, SVHN, and STL-10. The verifier behaves similarly across the three $L_\infty$ attacks, while CW $L_2$ varies across datasets. It also achieves substantially lower bypass rates than the baseline detectors.}
\label{fig:two_panel_results}

\end{figure*}

The proposed verifier achieves the strongest direct adaptive detection results across the three datasets and reported metrics in Table~\ref{tab:baseline_results}. Its AUC exceeds that of the strongest baseline on every dataset, while the directly attacked confidence-based and feature-based detectors have substantially higher bypass rates. The comparison isolates each detector's behavior when the attack is optimized against its own score.

\begin{table*}[t]
\caption{Direct full versus no-signature comparison under each detector-aware attack. Values are mean $\pm$ sample standard deviation in percent, over three seeds. Each evaluation retains 1,000 successful targeted examples.}
\label{tab:signature_control}
\centering
\scriptsize
\setlength{\tabcolsep}{5pt}
\begin{tabular}{llcccc}
\hline
Dataset & Attack & Full AUC $\uparrow$ & No-signature AUC $\uparrow$ & Full bypass $\downarrow$ & No-signature bypass $\downarrow$ \\
\hline
CIFAR-10 & PGD & $79.10{\pm}2.22$ & $83.24{\pm}0.50$ & $57.23{\pm}3.04$ & $51.07{\pm}2.30$ \\
CIFAR-10 & APGD & $79.06{\pm}2.21$ & $83.33{\pm}0.45$ & $57.50{\pm}2.95$ & $51.00{\pm}2.36$ \\
CIFAR-10 & AA+Refine & $79.16{\pm}2.19$ & $83.38{\pm}0.50$ & $57.10{\pm}2.93$ & $50.77{\pm}2.17$ \\
CIFAR-10 & CW $L_2$ & $64.54{\pm}1.39$ & $65.66{\pm}0.15$ & $95.63{\pm}1.17$ & $95.37{\pm}0.61$ \\
\hline
SVHN & PGD & $80.50{\pm}1.14$ & $85.54{\pm}0.50$ & $56.35{\pm}1.48$ & $53.15{\pm}2.19$ \\
SVHN & APGD & $80.35{\pm}1.02$ & $85.49{\pm}0.53$ & $56.10{\pm}0.57$ & $53.25{\pm}2.47$ \\
SVHN & AA+Refine & $80.98{\pm}1.03$ & $85.75{\pm}0.42$ & $55.20{\pm}1.13$ & $52.30{\pm}1.56$ \\
SVHN & CW $L_2$ & $70.54{\pm}0.08$ & $73.79{\pm}2.03$ & $95.10{\pm}0.85$ & $95.20{\pm}0.00$ \\
\hline
\end{tabular}
\end{table*}

Table~\ref{tab:signature_control} provides paired comparisons of the full verifier and a no-signature control. Both variants use the same verifier training, distillation, calibration, and rejection rule; the control sets $\lambda_{\mathrm{s}}=0$ instead of 0.05. Signature guidance improves clean verifier accuracy on CIFAR-10 and, to a smaller extent, SVHN. The no-signature variant obtains higher adaptive AUC across the paired attacks. These configurations also provide a two-point comparison of $\lambda_{\mathrm{s}}$ at fixed $\gamma=0.25$.

The CIFAR-10 component ablation further separates the contributions of the training objectives. Removing adaptive training reduces PGD AUC from 81.66\% to 0.98\% and increases bypass from 53.80\% to 99.80\%. Removing distillation preserves comparable AUC but reduces clean verifier accuracy.

The additional DenseNet121 experiment on CIFAR-10 exhibits the same clean-accuracy pattern as ResNet-18: signature guidance improves clean verifier accuracy, while the no-signature variant obtains higher adaptive AUC. The matched evaluation uses the same source images and target classes for both variants, with each attack optimized against its own verifier. A separate, larger PGD evaluation of the full DenseNet121 model provides an additional check.

\section{Discussion and Future Directions}
\label{sec:discussion}

The experiments suggest that effective verification benefits from separating the classifier's prediction from the evidence used to accept it. Detector-aware training teaches the verifier to respond to examples designed to challenge its rejection rule, rather than relying solely on the confidence or activation score of the fixed classifier. The consistent behavior across PGD, APGD, and AA+Refine further suggests that the learned verification signal captures properties shared by different $L_\infty$ attack procedures. This motivates further study of prediction-conditioned verification as a modular complement to existing classifiers.

The ablations also distinguish the objectives served by adaptive training, signature guidance, and distillation. Adaptive training is the primary source of the observed detection benefit, whereas signature guidance improves the verifier's clean predictions in the evaluated comparisons. The difference between the full and no-signature configurations suggests that clean predictive fidelity and adaptive detection should be considered jointly when choosing the signature weight. A natural extension is to investigate adaptive weighting schedules or class-dependent signature guidance, rather than using a single fixed coefficient. Distillation can be studied within the same framework as a way to preserve the fixed classifier's clean decision structure while the verifier learns from difficult examples.

The prediction-conditioned rejection rule provides a separate point of control. Class-specific thresholds allow the detector's operating point to reflect differences among predicted classes without requiring activation extraction at inference. Future work could examine alternative calibration objectives, including class-wise detection requirements and selective-classification criteria, to determine how the verifier's score can be used in applications with different acceptance policies. The training-only signature bank also keeps deployment independent of signature storage and comparison, making inference-cost studies and integration with existing classification pipelines relevant next steps.

Finally, the DenseNet121 experiment provides an initial indication that the clean-accuracy effect of signature guidance is not specific to ResNet-18. Broader studies could extend the paired evaluation to additional architectures, datasets, and training seeds. The evaluation with multiple attack objectives and perturbation norms also motivates a systematic study of multi-norm and multi-attack training curricula. Such evaluations would help characterize which aspects of the learned verification signal transfer across attack objectives and perturbation models.

\section*{Acknowledgment}
\textit{AI-use declaration:} OpenAI ChatGPT assisted with drafting and editing the Abstract and Sections~I--V. The authors take full responsibility for the manuscript.

\bibliographystyle{ieeetr}
\bibliography{references}

\end{document}